\documentclass[letterpaper]{article} 
\usepackage{aaai25}  
\usepackage{times}  
\usepackage{helvet}  
\usepackage{courier}  
\usepackage[hyphens]{url}  
\usepackage{graphicx} 
\usepackage{natbib}  
\usepackage{caption} 
\usepackage{algorithm}
\usepackage{algorithmic}

\usepackage{newfloat}
\usepackage{listings}
\DeclareCaptionStyle{ruled}{labelfont=normalfont,labelsep=colon,strut=off} 
\floatstyle{ruled}
\newfloat{listing}{tb}{lst}{}
\floatname{listing}{Listing}
\usepackage{amsmath}
\usepackage{booktabs}
\usepackage{multirow}
\usepackage{booktabs}
\usepackage{multirow}
\usepackage{diagbox}
\usepackage{tabulary}
\usepackage{comment}
\usepackage{colortbl}
\usepackage{tablefootnote}
\usepackage{tabularx}
\usepackage{soul}
\usepackage{subcaption}
\usepackage{overpic}
\usepackage{amsmath}
\usepackage{amssymb}
\usepackage{bm}
\usepackage{amsfonts}
\usepackage{mathtools}
\usepackage{subcaption}
\usepackage[dvipsnames]{xcolor}
\usepackage{xspace}
\usepackage{mathrsfs}

\usepackage{makecell}

\usepackage[symbol]{footmisc}

\newcommand{\ve}[1]{\mbox{{\bf #1}}} 

\newcommand{\boldtitle}[1]{{\noindent}{\textbf{#1}}.}

\newcommand{\onedot}{.\xspace}
\def\eg{\emph{e.g}\onedot} 

\def\ie{\emph{i.e}\onedot}

\newcommand{\SDFNet}{f_{s}} 
\newcommand{\UncNet}{f_{u}} 
\newcommand{\ColorNet}{f_{c}} 
\newcommand{\SdfFeat}{\mathscr{h}}
\newcommand{\NetParams}{\mathrm{\Phi}} 

\newcommand{\ViewDir}{\mathbf{v}}
\newcommand{\CamOrig}{\mathbf{o}}
\newcommand{\Pixel}{\mathbf{p}}
\newcommand{\normal}{\mathbf{n}}
\newcommand{\surfpts}{\mathcal{T}}
\newcommand{\plane}{\mathbf{\Pi}}

\newcommand{\image}{\mathbf{I}}
\newcommand{\patchset}{\ve{P}}

\definecolor{myPurple}{rgb}{0.4, .0, .8}
\definecolor{myGreen}{rgb}{0, 0.6, .3}
\definecolor{myRed}{rgb}{0.8, .2, .2}
\definecolor{myOrange}{rgb}{0.8, 0.45, 0.0}
\definecolor{myBlue}{rgb}{.0, .0, 1.0}
\definecolor{myBlue2}{rgb}{.0, 1.0, 1.0}
\definecolor{myBlack}{rgb}{.0, .0, 0.0}
\definecolor{darkmidnightblue}{rgb}{0.0, 0.2, 0.4}

\makeatletter
\newcommand{\printfnsymbol}[1]{%
        \textsuperscript{\@fnsymbol{#1}}%
}
\makeatother

\title{GURecon: Learning Detailed 3D Geometric Uncertainties \\ for Neural Surface Reconstruction}

\author{ 
    Zesong Yang\textsuperscript{\rm 1}\thanks{Contribute equally to this work.},
    Ru Zhang\textsuperscript{\rm 1}\printfnsymbol{1},
    Jiale Shi\textsuperscript{\rm 1}\printfnsymbol{1},
    Zixiang Ai\textsuperscript{\rm 1},
    Boming Zhao\textsuperscript{\rm 1}, \\
    Hujun Bao\textsuperscript{\rm 1},
    Luwei Yang\textsuperscript{\rm 2},
    Zhaopeng Cui\textsuperscript{\rm 1}\thanks{Corresponding author.} 
}
\affiliations{
    \textsuperscript{\rm 1}State Key Lab of CAD \& CG, Zhejiang University, \\
    \textsuperscript{\rm 2}Simon Fraser University \\
    zesongyang0@zju.edu.cn, ruzhang@zju.edu.cn, jlshi@zju.edu.cn, zxai@zju.edu.cn, 
    bmzhao@zju.edu.cn, bao@cad.zju.edu.cn, luweiy@sfu.ca, zhpcui@gmail.com
    \\
    \vspace{0.25 em}
    \small{Project Page: \url{https://zju3dv.github.io/GURecon/}}
}

\usepackage{bibentry}

\begin{document}
\maketitle

\begin{abstract}
Neural surface representation has demonstrated remarkable success in the areas of novel view synthesis and 3D reconstruction. However, assessing the geometric quality of 3D reconstructions in the absence of ground truth mesh remains a significant challenge, due to its rendering-based optimization process and entangled learning of appearance and geometry with photometric losses. 
In this paper, we present a novel framework, GURecon, which establishes a geometric uncertainty field for the neural surface based on geometric consistency.
Different from existing methods that rely on rendering-based measurement, 
GURecon models a continuous 3D uncertainty field for the reconstructed surface, and is learned by an online distillation approach without introducing real geometric information for supervision. 
Moreover, in order to mitigate the interference of illumination on geometric consistency, a decoupled field is learned and exploited to finetune the uncertainty field. 
Experiments on various datasets demonstrate the superiority of GURecon in modeling 3D geometric uncertainty, as well as its plug-and-play extension to various neural surface representations and improvement on downstream tasks such as incremental reconstruction. 
\end{abstract}

\section{Introduction}
\label{sec:intro}
Image-based 3D reconstruction is a long-standing problem in computer vision with a wide range of applications like AR/VR, autonomous driving, digital heritage preservation, etc. 
Recently, learning-based methods have attracted much attention with the development of neural radiance representations like Neural Radiance Fields (NeRF) \cite{mildenhall2021nerf}. Unlike traditional methods, NeRF and its variants (\eg, NSVF~\cite{liu2020neural}, NeuS~\cite{wang2021neus}) encode scene geometry and appearance with neural networks, which can be optimized by leveraging the differentiable rendering given a set of calibrated images. 

Although the neural representations demonstrate remarkable performance in novel view synthesis and surface reconstruction with high levels of detail and photorealism, assessing the reconstruction quality remains challenging. Some existing work incorporates uncertainty estimation into NeRF models to identify areas with poor rendering quality.
NeRF-W \cite{martin2021nerf} and its following works \cite{pan2022activenerf, ran2023neurar} take the radiance field as Gaussian distributions to model the uncertainty of rendered RGB.  
Some other works model uncertainty as the entropy of the weight distribution along rays in NeRF models \cite{zhan2022activermap,lee2022uncertainty}. Besides, the deep learning techniques are also applied to NeRF to quantify the uncertainty via ensemble learning \cite{sunderhauf2023density} or variational inference \cite{shen2021stochastic, shen2022conditional}. 
However, all these methods evaluate the uncertainty of the neural fields in a single-view pixel-wise manner via volumetric rendering, which does not support direct evaluation of 3D geometry accurately, and the uncertainty for the same surface point may vary across different views due to the multi-view inconsistencies in images caused by varying lighting and observation angles, disobeying the view-independent nature of 3D geometric uncertainties. 

\begin{figure}[t] \centering
    \includegraphics[width= \linewidth]{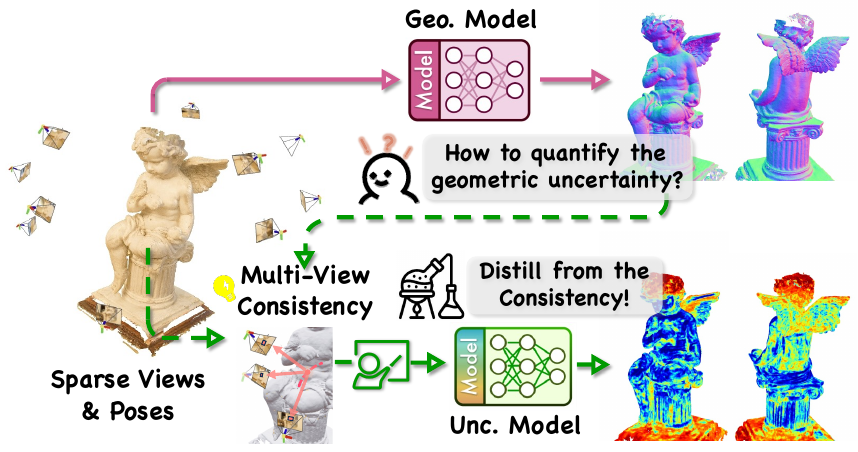}
    \caption{
    \textbf{A brief overview.} 
    By leveraging multi-view consistency as guidance, GURecon learns detailed 3D geometric uncertainties for neural surface reconstruction.
} 
\label{fig:teaser}
\end{figure}

In this paper, we present a novel framework, \ie, GURecon, which is able to learn detailed 3D geometric uncertainty for neural surface reconstruction as shown in Fig.~\ref{fig:teaser}. Different from existing methods relying on rendering-based pixel-wise uncertainty measurement, GURecon directly models the 3D uncertainty for surface points and ensures consistency over viewpoints.
However, designing such a system is nontrivial. Without ground truth geometric supervision (\eg, input depth), it is difficult to model the geometric uncertainty just based on photometric error between the rendered and input images. This is because, as with previous methods, the neural radiance field tends to overfit the input images in sparse viewpoint settings, resulting in minor photometric errors but significant geometric errors, just as the ambiguity problem highlighted in NeRF++~\cite{zhang2020nerf++}.

Motivated by the traditional multi-view stereo~\cite{hu2012quantitative,schonberger2016pixelwise} where photometric consistency is widely used to assess the confidence of reconstructed geometry, we employ the multi-view consistency as a cue to quantify the quality of reconstruction.
We compute the patch-based warping consistency of surface points projected onto the input images, and utilize it as a pseudo label of geometric accuracy to supervise a continuous geometric uncertainty field based on a novel online distillation approach. 
We consider the estimated uncertainty derived from such pseudo label as epistemic uncertainty, which reflects the geometric confidence of the model per-scene trained with given images (\ie, reconstruction error), and serves as a reference identifying areas where reconstruction is inadequate and unreliable. 

Besides, inevitable illumination in real-world scenes poses a challenge to modeling geometric uncertainty based on inconsistent color observations. To handle this problem, we propose to learn additional decoupled fields and further fine-tune the uncertainty field by removing view-dependent factors from each image. 
Our method can be extended to various neural surface representations. 
With accurate 3D geometric uncertainty estimation, GURecon can be integrated into tasks like incremental reconstruction to boost the quality of surface reconstruction. 

Our main contributions are summarized as follows:
\begin{itemize}
    \item We present a novel framework, \ie, GURecon, to quantify geometric uncertainty for neural surface reconstruction. 
    \item We proposed a new strategy to distill geometric uncertainty based on multi-view consistency, thus decoupling geometric uncertainty with rendering-related uncertainty.
    \item Additional decoupled fields are learned and exploited to eliminate view-dependent factors for robust estimation.
    \item Extensive experiments on diverse datasets demonstrate the superior performance of our framework in modeling geometric uncertainty and the potential for application in downstream tasks such as incremental reconstruction.
\end{itemize}

\section{Related Work}
\noindent\textbf{Neural Surface Reconstruction.}
Neural representations have achieved great success in various tasks such as multi-view 3D reconstructions and novel view synthesis. Among them, NeRF~\cite{mildenhall2021nerf} encodes scenes within an MLP through differentiable volume rendering, enabling high-quality novel view synthesis. 
SDF-based variants~\cite{wang2021neus,yariv2021volume} constrain the scene as an SDF field and achieve smooth surface reconstruction. Subsequent works utilize monocular geometric priors~\cite{yu2022monosdf,xiao2024debsdf} and geometric consistency to enhance the quality of reconstruction. \cite{fu2022geo,darmon2022improving} utilize the homography warp as a constraint, while \cite{ge2023ref,wang2022neuris} use multi-view consistency to filter the interferences in input data.
In this paper, we use a hash-based NeuS~\cite{zhao2022human} as the scene representation and first utilize multi-view consistency as guidance for uncertainty quantification.

\noindent\textbf{Uncertainty Modeling in NeRF.}
Considering the various interferences such as dynamic objects and limited observations present in input data, integrating uncertainty modeling becomes crucial for achieving robust reconstructions.
Uncertainty estimation in NeRF can be divided into epistemic uncertainty and aleatoric uncertainty. The former typically arises from data limitations, while the latter is generally associated with the inherent randomness of data. NeRF-Wild~\cite{martin2021nerf} mitigate the interference of transient objects by modeling rendered colors as Gaussian distributions. Subsequent works build upon it to address the Next Best View (NBV) problem~\cite{pan2022activenerf, chen2023leveraging}. 
Other approaches tackle uncertainty through sampling techniques to establish a probability model, such as ensemble learning~\cite{sunderhauf2023density} or variational inference~\cite{shen2021stochastic,shen2022conditional}, the former is time and memory-consuming, while the latter involves major network architecture modifications.
In contrast to predicting probability, \cite{lee2022uncertainty} computes uncertainty as the entropy of weight distribution along the rays. All these methods utilize probabilistic models to model uncertainty, focusing on network convergence rather than constructing uncertainty from a geometric perspective.
Bayes' Rays~\cite{goli2023bayes} simulates spatially parameterized perturbation of the radiance field and uses a Laplace approximation to produce a volumetric uncertainty field. Another work, FisherRF~\cite{jiang2023fisherrf}, introduces fisher information for uncertainty modeling. However, they still model uncertainty in a pixel-wise manner based on rendering RGB values and need to measure uncertainty by rendering at a pixel level, not approaching the problem from a 3D geometric perspective.
All existing methods are designed for uncertainty estimation in NeRF, considering only the rendering perspective, with no work addressing geometric uncertainty estimation for neural surface representation. In contrast, we introduce GURecon, the first framework that models geometric uncertainty for the neural surface from the perspective of multi-view geometric consistency.

\section{Method}
\label{sec:method}
\begin{figure*}[t]
	\centering
	\scriptsize
	\includegraphics[width=\textwidth]{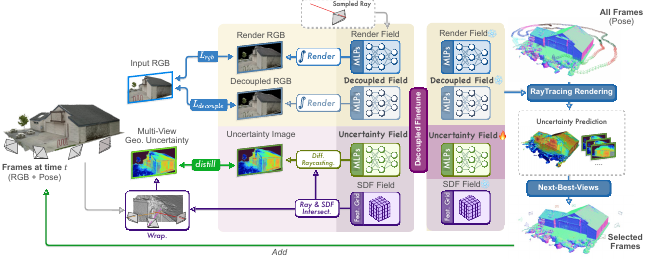}
	\caption{\textbf{System Overview.} 
 The proposed GURecon models a geometric uncertainty field supervised by the pseudo labels computed based on the multi-view geometry consistency.
 To deal with the view-dependent factors,  additional decoupled fields are also learned and exploited to fine-tune the uncertainty field.  With the predicted uncertainty fields, GURecon can boost the downstream tasks such as incremental reconstruction. 
 }
\label{fig:pipeline}
\end{figure*}

In this paper, we introduce a novel framework, \ie, GURecon, which enables accurate geometric uncertainty estimation for various neural surface representations without GT geometric information for supervision. As shown in Fig.~\ref{fig:pipeline}, with given posed images, we learn a render field and an SDF field through differentiable rendering.
As the training progresses, with the currently learned geometry field, we first utilize a root-finding method to identify the zero-crossing points intersected with the implicit surface and calculate the multi-view consistency of these points as pseudo supervision to guide the learning of geometric uncertainty (Sec.~\ref{sec:patch_MVC}). 
Then we present a novel online distillation method that simultaneously learns a spatially continuous uncertainty field with other fields in a self-supervised manner by utilizing the multi-view consistency as pseudo ground truth labels (Sec.~\ref{sec:distillation}). 
In order to overcome the interference caused by view-dependent factors in the calculation of multi-view consistency, we propose to simultaneously learn additional decoupled fields and exploit them to fine-tune the geometric uncertainty field (Sec.~\ref{sec:decouple}).

\subsection{Neural Surface Representation}
Taking NeuS~\cite{wang2021neus} as a representative, we denote each pixel by casting a ray as $ \Pixel(t)= \CamOrig+t\ViewDir$, where $\CamOrig$ is the camera origin and $\ViewDir$ is the view direction. 
We define the surface as the zero-level set $\mathcal{S}=\{ x\in \mathbb{R}^{3}|\SDFNet(x)=0 \}$ using a geometry encoder $\SDFNet(x;\NetParams_{s})$, which predicts the signed distance field (SDF) value $s$ and a hidden geometric feature $\SdfFeat$ at point $x$. Additionally, we employ a radiance encoder $\ColorNet(x, \ViewDir, \normal, \SdfFeat;\NetParams_c)$ to predict the color $c$ based on the view direction $\ViewDir$, where $\normal$ is the normal at point $x$ computed from the gradient of the SDF. The color of each pixel is computed by accumulating colors of sampled points along the ray:
\begin{equation}
\begin{split}
    \hat{C}(\Pixel) &= \sum_{i=1}^{N} T_i \alpha_i {\mathbf{c}}_i, \;
    T_i = \prod_{j=1}^{i-1}(1-{\alpha}_j), \\
    {\alpha}_j &= \max \left (\frac{\psi_s(s_{i}) - \psi_s(s_{i+1})}{\psi_s(s_i)}, 0 \right),
\end{split}
\end{equation}
where $T_i$ is accumulated transmittance at $\Pixel(t_i)$, $\psi_s$ is the sigmoid function, and $\alpha_i$ is the opacity of the $i$-th ray segment. Similar neural surface representations can also be adopted as long as the geometric surface can be computed on the fly.

\subsection{Patch-based Multi-view Consistency}
\label{sec:patch_MVC}
Motivated by the traditional MVS works \cite{stereopsis2010accurate,schonberger2016pixelwise} that leverage photometric consistency among different views as a geometric constraint, we exploit it as a cue to guide the learning of geometric uncertainty in the neural surface.

\noindent\textbf{Surface Interaction Retrieval.} The primary step is to identify the surface points of the neural representation. Following 
the existing works~\cite{fu2022geo,oechsle2021unisurf}, root finding is a widely used method to locate the intersection with the neural surface. As our approach is based on SDF representation, and the SDF values of sampling points along the ray are precomputed for volume rendering,  we employ linear interpolation to locate the zero-crossing points $\surfpts$ as follows:
\hspace*{-0.5em} 
\begin{equation}
\begin{split}
\mathcal{T} &= \left\{
\Pixel(t_i^*) \mid t_i^* = \frac{\hat{s}_i \hat{t}_{i+1} - \hat{s}_{i+1} \hat{t}_i}
                            {\hat{s}_i - \hat{s}_{i+1}}
    \right\}, \\
\hat{t}_i &= \underset{i}{\arg \min} \left\{ t_i \mid s_i \cdot s_{i+1} < 0 \right\},
\end{split}
\end{equation}
where $s_i$ is the SDF value of $\Pixel(t_i)$, \ie, $\SDFNet(\Pixel(t_i))$, and $\hat{t}_i$ is the ray segment of the zero-crossing point.

\noindent\textbf{Patch-based Multi-view Photometric Consistency.} 
With the intersected points $\surfpts$ of the neural surface, we acquire the multi-view photometric information by projecting these points onto visible views following \cite{fu2022geo,darmon2022improving}. 
For robustness, we consider the consistency of the pixel patches around the projection of surface points rather than a single pixel. We approximate the small region around the point as a local plane and use the homography warp to compute the patch-based multi-view photometric consistency for computational efficiency. The tangent plane \cite{stereopsis2010accurate,schonberger2016pixelwise} at the surface point $\Pixel^{\prime}$ can be modeled as follows:
\begin{equation} 
\label{eq:plane_repre}
\plane=\{\normal^{\prime}, \Pixel^{\prime} \ |\ {\normal^{\prime}}^T\Pixel^{\prime}+d=0\},\; {\rm where} \ \Pixel^{\prime} \in \surfpts,
\end{equation}
where $\normal^{\prime}$ is the normal computed from the gradient of the SDF values, \ie, $\nabla \SDFNet(\Pixel^{\prime})$, and $d$ is the distance to the origin of the coordinate system.  Then, the homography warping matrix $\ve{H}$ can be constructed based on the local plane and enables the mutual projection of image patches between viewpoints as the following:
\begin{equation} 
\label{homowarp}
\ve{H}_{rel}=\ve{K}_{src}\left(\ve{R}_{rel}-\mathbf{t}_{rel}
\frac{ {\normal^{\prime}}^T}{d}\right) \ve{K}_{ref}^{-1}, \; 
\ve{P}_j=\ve{H}_{ij}\ve{P}_i,
\end{equation}
where \ve{K} corresponds to the camera's intrinsic matrix, [$\ve{R}_{rel}$,$\ve{t}_{rel}$] corresponds to the relative transformation matrix from the reference view $i$ to the source view $j$, 
$\ve{P}_i$ and $\ve{P}_j$ represent the corresponding patch coordinates of the local plane projected on reference and source view respectively.

\begin{figure}[t] \centering
    \includegraphics[width=\linewidth]{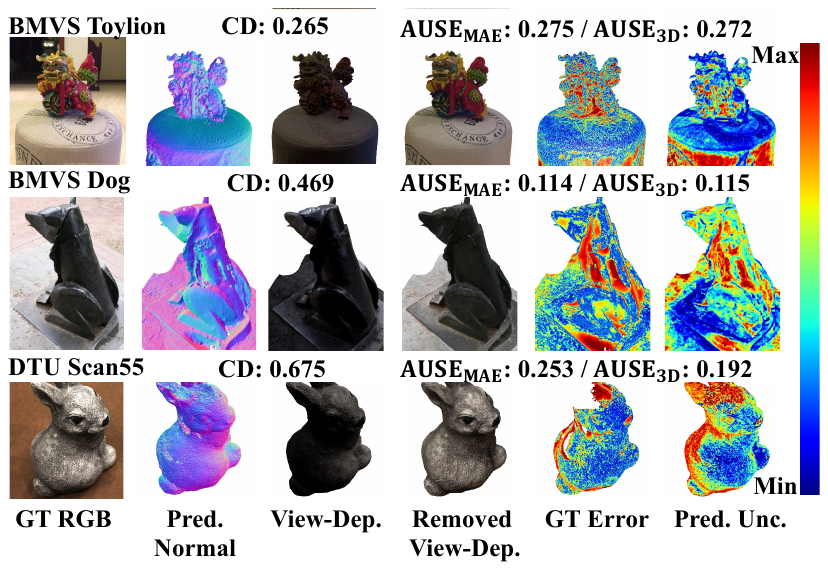}
    \caption{\textbf{Visualization of the learned fields.}
    Our method presents accurate decoupled results for view-dependent factors, and the learned geometric uncertainties are well aligned with the GT geometric error.
    } 
    \label{fig:ours_res}
\end{figure}

Finally, we convert the color images {$\image_i$} into gray images {$\image^{\prime}_i$}, and utilize the Structural Similarity Index Measure~(SSIM)~\cite{campbell2008using} to measure the correlation coefficient $\mathbb{C}$ between pairs of projected patches as:
\begin{equation} 
\label{Multi-view Consistency}
\mathbb{C}_{ij}^k =1- SSIM(\image^{{\prime}}_i(\patchset_i^k),\image^{\prime}_j(\patchset_j^k)).
\end{equation}
As the similarity between $\image^{{\prime}}_i(\patchset_i^k)$ and $\image^{\prime}_j(\patchset_j^k)$ increases, the score of $\mathbb{C}_{ij}^k$ decreases and the corresponding geometric quality reconstructs better. Considering the potential occlusion and large deviations in projection viewing angles, for robustness we ultimately select the four patch pairs with the lowest computed scores 
$\mathbb{C}_n^{k*}$ and compute the average score to represent the final geometric consistency ${\mathbb{G}_k}$ of $\Pixel_k$: 
\begin{equation}
\mathbb{G}_k= (\textstyle\sum_{n=1}^4{\mathbb{C}_n^{k*}})/4, \; 
{\rm where} \; 
\mathbb{C}_n^{k*} \in \overset{4}{\underset{i, j}{\operatorname{argmin}}}\left\{\mathbb{C}_{i j}^k\right\}.
\end{equation}
We utilize the computed consistency as pseudo-ground-truth labels to guide the learning of geometric uncertainty.

\subsection{Distillation of Geometric Uncertainty Field}
\label{sec:distillation}
Considering it is impractical and inefficient to perform such consistency calculations for each pixel during inference due to its high computational cost, we propose to learn a geometric uncertainty field distilled from the above geometric consistency, which is conducted simultaneously with the learning process of geometric and radiance fields.

Specifically, considering that geometric uncertainty is a view-independent factor which only related to the position $x$ of points, we use an uncertainty field $\UncNet(x;\NetParams_u)$ with position input and learn the geometric uncertainty solely for surface points $\Pixel^{\prime} \in \surfpts$ corresponding to the current SDF field during training process.

As described in Sec~\ref{sec:patch_MVC}, we firstly use the root-finding method to locate the surface point corresponding to the current iteration at each training step, and then take the multi-view patch-based consistency of the point as a pseudo label to supervise a continuous and accurate uncertainty field using online distillation with the following loss:
\begin{equation} 
\label{online distillation}
\mathcal{L}_{distill}=\frac{1}{ \mathcal{R^{\prime}} }\displaystyle\sum_{r\in \mathcal{R^{\prime}}}\left | \UncNet({\Pixel_{r}}^{\prime})-\mathbb{G}_r\right |, 
\end{equation}
where $\mathcal{R^{\prime}}$ corresponds to the set of rays intersected with the surface, and ${\Pixel_{r}}^{\prime}$ is the intersection sampled on ray $r$.

\begin{figure}[t]
	\centering
	\includegraphics[width=0.475\textwidth]{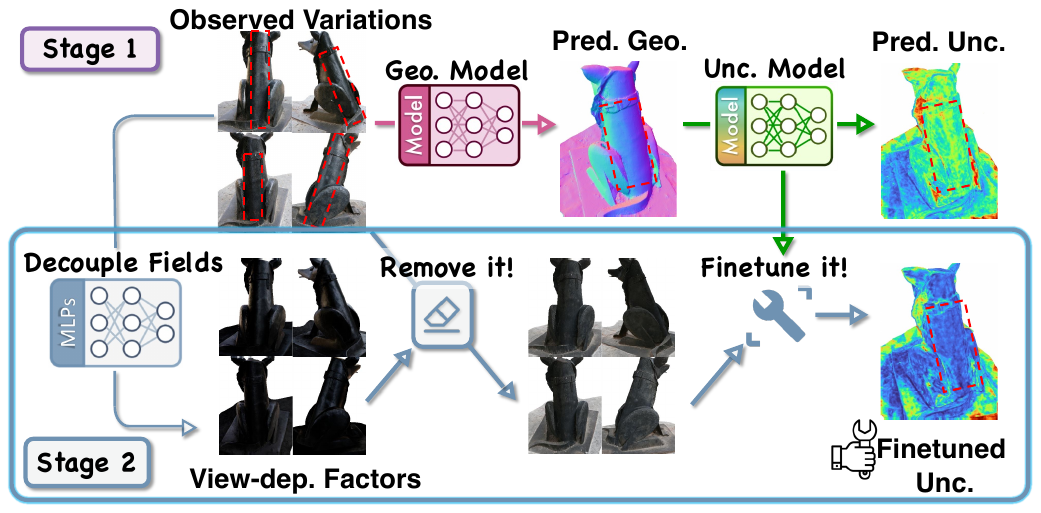}
    \caption{\textbf{Uncertainty finetuning with the decoupled fields.}
    Even if the geometry has been well reconstructed, the uncertainty field still erroneously estimates it with high uncertainty caused by light interference across different views.
    We employ the decoupled fields to remove the view-dependent factor from the training images. 
    }
\label{fig:finetune}
\end{figure}

\subsection{Finetuning with Decoupled Fields}
\label{sec:decouple}
Variations in light across different views can lead to inconsistent observations and subsequently impact the computation of multi-view consistency as Fig.~\ref{fig:finetune} shows.
For more accurate modeling of the geometric uncertainty, inspired by prior works for radiance decomposition \cite{verbin2022refnerf,fan2023factored,tang2023delicate}, we further introduce an additional branch to decouple view-dependent factors as:
\begin{equation} 
\label{decouple branch}
\bar{C}=C_{vi}(\normal^{\prime},\ \Pixel^{\prime},\ \SdfFeat^{\prime})\ + \ C_{vd}(\normal^{\prime},\ \Pixel^{\prime},\ {{w}_{r}}^{\prime},\  \SdfFeat^{\prime}),
\end{equation}
where $C_{vd}$ and $C_{vi}$ correspond to view-dependent and view-independent components respectively. As the same with uncertainty field, we only decouple the points on the surface, and ${{w}_{r}}^{\prime}$ is the reflection of the view direction around the normal $\normal^{\prime}$. 
We model view-dependent components using the reflection direction rather than the view direction as it allows for better interpolation of factors like specular following \cite{ge2023ref,fan2023factored,verbin2022refnerf}. The decouple fields are trained with surface rendering as follows:
\begin{equation} 
\label{decouple branch}
\mathcal{L}_{decouple}=\frac{1}{\left | \mathcal{R^{\prime}}\right |}\displaystyle\sum_{r\in \mathcal{R^{\prime}}}\left |
{\bar{C}}^r-C_{gt}^r
 \right |.
\end{equation}

Once the fields are decoupled, we eliminate lighting, reflections, and other interferences by subtracting the rendered view-dependent factor $\image_{vd}$ from the true RGB image as shown in Fig.~\ref{fig:ours_res}, and then use the processed image to recompute the multi-view consistency as the same in Sec.~\ref{sec:patch_MVC}, thus fine-tuning the uncertainty field. 
We divided our pipeline into two stages. In the first stage we sample on the ground-truth image to generate pseudo labels and supervise the geometric uncertainty while simultaneously learning the decoupled fields, and in the second stage, our method initially renders the view-dependent component for each training viewpoint and processes the ground-truth image, then freezes other fields and uses the processed image for $N_{ft}$ iterations to finetune the uncertainty field as shown in Fig.~\ref{fig:finetune}.

\begin{figure}[t] 
    \centering
    \includegraphics[width=0.45\textwidth]{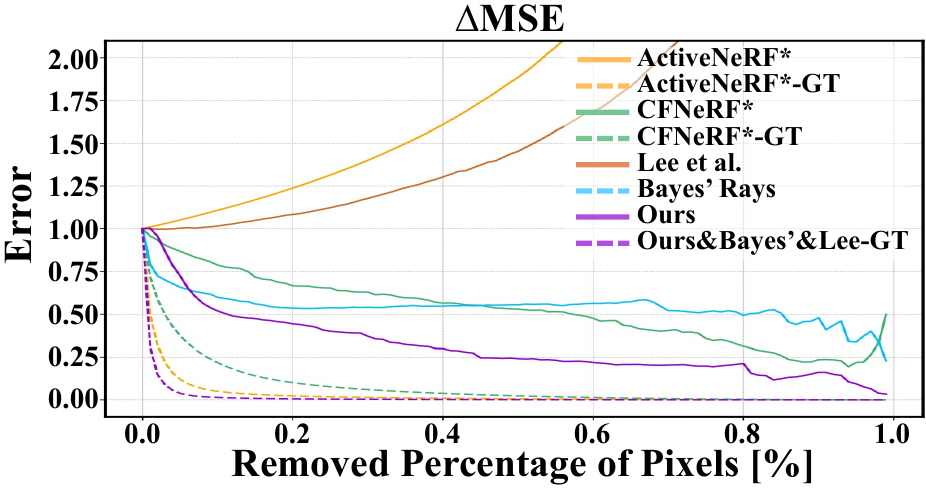} 
    \caption{\textbf{Sparsification curves of different methods.} The dashed and solid lines correspond to the average error of the remaining pixels filtered using GT-error-based and uncertainty-based criteria, the area between them is AUSE. Bayes' Rays and Lee et al.~\cite{lee2022uncertainty} share the same GT curve with ours as post-hoc frameworks.
    }
    \label{fig:ause_curve}
\end{figure}

\subsection{Loss Function and Implementation Details}
\boldtitle{Loss Function}  Our total loss is defined as the following:
\begin{equation} 
\label{total loss}
\begin{split}
\mathcal{L} = &\mathcal{L}_{color} + \alpha_1 \mathcal{L}_{reg} + \alpha_2 \mathcal{L}_{mask}  + \\ 
&\alpha_3\mathcal{L}_{decouple} + \alpha_4\mathcal{L}_{distill}.
\end{split}
\end{equation}
Following the definition in NeuS~\cite{wang2021neus},
$\mathcal{L}_{color}$ is the rgb loss between the ground truth pixel colors and the rendered colors, and $\mathcal{L}_{reg}$ is the eikonal loss to regularize the gradients of SDF.
Since we only focus on the geometric quality of the target to be reconstructed, we use the mask to filter irrelevant regions and $\mathcal{L}_{mask}$ corresponds to the constraint. We set $\alpha_1=0.1$, $\alpha_2=1.0$, $\alpha_3=0.1$ and $\alpha_4=0.1$.

\boldtitle{Implementation Details} 
The GURecon serves as a plug-and-play module applicable to various neural surface representations. Our preference for the fundamental 3D representation leans towards the hash-based variant \cite{zhao2022human} due to its time efficiency. 
For each scene, we sample 1024 rays per batch and train for 50k iterations, which takes nearly $30$ minutes on an NVIDIA RTX 3090. 
After completing the training stage, we run an additional 10k iterations to finetune the uncertainty field while keeping other fields frozen. Since the geometry is fixed, we utilize sphere tracing instead of inefficient sampling to locate the intersection points with the neural surface, and the fine-tuning stage takes approximately 5 minutes. Please refer to the supplementary materials for more details.

\begin{table}[tb] 
    \centering
    \resizebox{\linewidth}{!}{
        \Huge 
        \begin{tabular}{lcccccc}
            \specialrule{.2em}{.1em}{.1em}
                \multirow{2}{*}{\diagbox{Scenes}{Methods}} & 
                \multicolumn{2}{c}{\textbf{ActiveNeRF\textsuperscript{*}}} & 
                \multicolumn{2}{c}{\textbf{Bayes' Rays}} & 
                \multicolumn{2}{c}{\textbf{Ours}} \\
                
            \cmidrule[0.5pt](rl){2-3} \cmidrule[0.5pt](rl){4-5} \cmidrule[0.5pt](rl){6-7} & 
            
            $\mathrm{AUSE}_\mathrm{3D}$($\downarrow$) & 
            CD($\downarrow$) & 
            ${\mathrm{AUSE}}_\mathrm{3D}$($\downarrow$) & 
            CD($\downarrow$) & 
            $\mathrm{AUSE}_\mathrm{3D}$($\downarrow$) & 
            CD($\downarrow$) \\
            
            \toprule
            
            TNT-Barn (100 images) & 1.076 & 1.079 & 0.438 & 0.994 & \textbf{0.327} & 0.994 \\
            
            TNT-Truck (65 images) & 0.989 & 3.082 & 0.289 & 2.965 & \textbf{0.243} & 2.965 \\
            
            TNT-Caterpillar (100 images) & 1.247 & 0.808 & 0.346 & 0.747 & \textbf{0.198} & 0.747 \\
            
            \specialrule{.2em}{.1em}{.1em}
            
        \end{tabular}
    }
    \caption{Uncertainty Quantification for TNT dataset.} 
    \label{tab:tnt_uc}
\end{table}

\begin{figure}[t] 
\centering
    \includegraphics[width=0.975\linewidth]{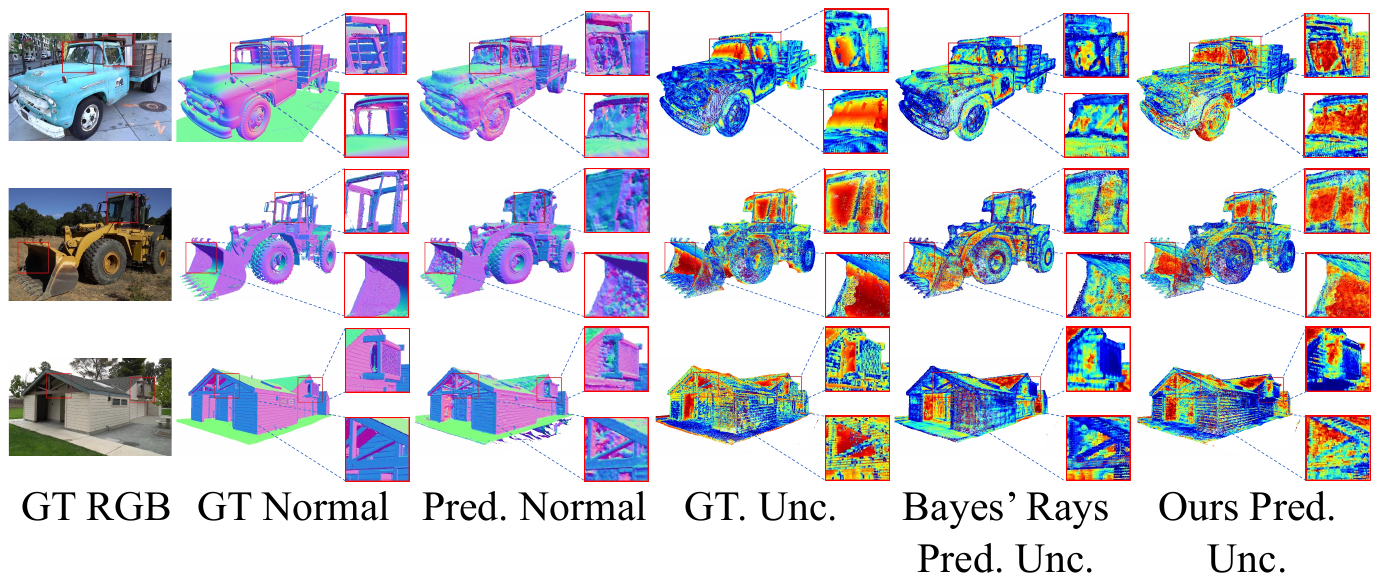}
    \caption{\textbf{Geometric uncertainty in TNT dataset.}
    We present the reconstruction results and the corresponding GT 3D error. GURecon predicts more accurate uncertainty than Bayes' Rays \cite{goli2023bayes}, especially in areas with texture repetition and reflection. 
    } 
    \label{fig:ours&bayes}
\end{figure}

\begin{table*}[t] 
    \centering
    \begin{subtable}{\linewidth}
        \centering
        \label{tab:bmvs_uc}
        \scalebox{0.7}
        {
            \begin{tabular}{lcccccccccccccccc}  
            
                \specialrule{.2em}{.1em}{.1em}
                
                    \multicolumn{1}{c}{Scenes} & 
                    \multicolumn{4}{c}{\textbf{BMVS-Angel}} & 
                    \multicolumn{4}{c}{\textbf{BMVS-Dog}} & 
                    \multicolumn{4}{c}{\textbf{BMVS-Egg}} & 
                    \multicolumn{4}{c}{\textbf{BMVS-Jade}} \\
                \cmidrule[0.5pt](rl){2-5}
                \cmidrule[0.5pt](rl){6-9}
                \cmidrule[0.5pt](rl){10-13}
                \cmidrule[0.5pt](rl){14-17}
        
                    \multirow{2}{*}{\diagbox{Methods}{Metrics}} & 
                    \multicolumn{3}{c}{AUSE($\downarrow$)} & 
                    \multirow{2}{*}{CD($\downarrow$)} & 
                    \multicolumn{3}{c}{AUSE($\downarrow$)} & 
                    \multirow{2}{*}{CD($\downarrow$)} & 
                    \multicolumn{3}{c}{AUSE($\downarrow$)} & 
                    \multirow{2}{*}{CD($\downarrow$)} & 
                    \multicolumn{3}{c}{AUSE($\downarrow$)} & 
                    \multirow{2}{*}{CD($\downarrow$)} \\
                \cmidrule[0.3pt](rl){2-4}                       
                \cmidrule[0.3pt](rl){6-8}                        
                \cmidrule[0.3pt](rl){10-12}                     
                \cmidrule[0.3pt](rl){14-16} & 
                    
                    $\Delta$MSE & $\Delta$MAE & 3D &  & 
                    $\Delta$MSE & $\Delta$MAE & 3D &  & 
                    $\Delta$MSE & $\Delta$MAE & 3D &  & 
                    $\Delta$MSE & $\Delta$MAE & 3D &  \\
        
                \toprule
        
                    CFNeRF\textsuperscript{*}  & 
                    \cellcolor[HTML]{FAC791}{0.384} & \cellcolor[HTML]{FAC791}{0.318} & \diagbox{}{} &  1.324 & 
                    0.354 & 0.287 & \diagbox{}{} & 1.904 & 
                    0.337 & 0.306 & \diagbox{}{} & 1.841 & 
                    0.335 & 0.291 & \diagbox{}{} & 1.433 \\
                    
                    ActiveNeRF\textsuperscript{*}  & 
                    1.425 & 1.158 & 1.471 & \cellcolor[HTML]{FAC791}{0.385} & 
                    0.796 & 0.686 & 0.786 & \cellcolor[HTML]{FAC791}{0.632} & 
                    0.608 & 0.631 & 0.606 & \cellcolor[HTML]{FAC791}{1.548} & 
                    1.513 & 0.946 & 1.067 & \cellcolor[HTML]{FAC791}{0.912} \\
        
                \hline
        
                    Lee et al.\textsuperscript{*}  & 
                    1.945 & 1.380 & \diagbox{}{} & \cellcolor[HTML]{F59194} & 
                    1.929 & 1.599 & \diagbox{}{} & \cellcolor[HTML]{F59194} & 
                    2.644 & 1.500 & \diagbox{}{} & \cellcolor[HTML]{F59194} & 
                    1.782 & 1.615 & \diagbox{}{} & \cellcolor[HTML]{F59194} \\
                    
                    Bayes' Rays  & 
                    0.521 & 0.469 & \cellcolor[HTML]{FAC791}{0.201} & \cellcolor[HTML]{F59194}{0.365$^\dagger$} & 
                    \cellcolor[HTML]{FAC791}{0.258} & \cellcolor[HTML]{FAC791}{0.225} & \cellcolor[HTML]{FAC791}{0.157} & \cellcolor[HTML]{F59194}{0.469$^\dagger$} & 
                    \cellcolor[HTML]{F59194}{0.147} & \cellcolor[HTML]{F59194}{0.172} & \cellcolor[HTML]{FAC791}{0.078} & \cellcolor[HTML]{F59194}{0.875$^\dagger$} & 
                    \cellcolor[HTML]{FAC791}{0.233} & \cellcolor[HTML]{FAC791}{0.238} & \cellcolor[HTML]{FAC791}{0.143} & \cellcolor[HTML]{F59194}{0.904$^\dagger$} \\
                    Ours  & 
                    \cellcolor[HTML]{F59194}{0.295} & \cellcolor[HTML]{F59194}{0.271} & \cellcolor[HTML]{F59194}{0.111} &  \cellcolor[HTML]{F59194} & 
                    \cellcolor[HTML]{F59194}{0.112} & \cellcolor[HTML]{F59194}{0.114} & \cellcolor[HTML]{F59194}{0.115} &  \cellcolor[HTML]{F59194} & 
                    \cellcolor[HTML]{FAC791}{0.224} & \cellcolor[HTML]{FAC791}{0.206} & \cellcolor[HTML]{F59194}{0.069} &  \cellcolor[HTML]{F59194} & 
                    \cellcolor[HTML]{F59194}{0.130} & \cellcolor[HTML]{F59194}{0.207} & \cellcolor[HTML]{F59194}{0.129} &  \cellcolor[HTML]{F59194} \\

                \specialrule{.2em}{.1em}{.1em}
                    \multicolumn{1}{c}{Scenes} & 
                    \multicolumn{4}{c}{\textbf{BMVS-Sculpture}} & 
                    \multicolumn{4}{c}{\textbf{BMVS-Soilder}} & 
                    \multicolumn{4}{c}{\textbf{BMVS-Stonelion}} & 
                    \multicolumn{4}{c}{\textbf{BMVS-Toylion}} \\
                \cmidrule[0.5pt](rl){2-5}                        
                \cmidrule[0.5pt](rl){6-9}                        
                \cmidrule[0.5pt](rl){10-13}                        
                \cmidrule[0.5pt](rl){14-17}
                    
                    \multirow{2}{*}{\diagbox{Methods}{Metrics}} & 
                    \multicolumn{3}{c}{AUSE($\downarrow$)} & 
                    \multirow{2}{*}{CD($\downarrow$)} & 
                    \multicolumn{3}{c}{AUSE($\downarrow$)} & 
                    \multirow{2}{*}{(CD$\downarrow$)} & 
                    \multicolumn{3}{c}{AUSE($\downarrow$)} & 
                    \multirow{2}{*}{CD($\downarrow$)} & 
                    \multicolumn{3}{c}{AUSE($\downarrow$)} & 
                    \multirow{2}{*}{CD($\downarrow$)} \\
                \cmidrule[0.3pt](rl){2-4}                        
                \cmidrule[0.3pt](rl){6-8}                        
                \cmidrule[0.3pt](rl){10-12}                      
                \cmidrule[0.3pt](rl){14-16} &
                
                    $\Delta$MSE & $\Delta$MAE & 3D &  & 
                    $\Delta$MSE & $\Delta$MAE & 3D &  & 
                    $\Delta$MSE & $\Delta$MAE & 3D &  & 
                    $\Delta$MSE & $\Delta$MAE & 3D &  \\
                    
                \toprule
                
                    CFNeRF\textsuperscript{*}   &
                    \cellcolor[HTML]{FAC791}{0.348} & \cellcolor[HTML]{FAC791}{0.399} & \diagbox{}{} & 1.346 & 
                    0.395 & 0.406 & \diagbox{}{} & 2.053 & 
                    0.463 & 0.495 & \diagbox{}{} & 1.712 &
                    0.372 & 0.315 & \diagbox{}{} & 1.987 \\
                    
                    ActiveNeRF\textsuperscript{*}  & 
                    1.270 & 1.056 & 1.279 & \cellcolor[HTML]{FAC791}{0.572} &
                    0.704 & 1.103 & 1.196 & \cellcolor[HTML]{FAC791}{0.569} & 
                    1.237 & 1.085 & 1.158 & \cellcolor[HTML]{FAC791}{0.493} & 
                    0.926 & 1.172 & 0.852 & \cellcolor[HTML]{FAC791}{0.462} \\
                    
                \hline
                    Lee et al.\textsuperscript{*}  & 
                    2.129 & 1.988 & \diagbox{}{} & \cellcolor[HTML]{F59194} & 
                    1.696 & 1.070 & \diagbox{}{} & \cellcolor[HTML]{F59194} & 
                    2.282 & 1.640 & \diagbox{}{} & \cellcolor[HTML]{F59194} & 
                    2.329 & 1.720 & \diagbox{}{} & \cellcolor[HTML]{F59194} \\
                    
                    Bayes' Rays  & 
                    0.720 & 0.564 & \cellcolor[HTML]{FAC791}{0.259} & \cellcolor[HTML]{F59194}{0.560$^\dagger$} & 
                    \cellcolor[HTML]{FAC791}{0.147} & \cellcolor[HTML]{FAC791}{0.195} & \cellcolor[HTML]{FAC791}{0.081} & \cellcolor[HTML]{F59194}{0.541$^\dagger$} &
                    \cellcolor[HTML]{FAC791}{0.296} & \cellcolor[HTML]{FAC791}{0.357} & \cellcolor[HTML]{FAC791}{0.223} & \cellcolor[HTML]{F59194}{0.477$^\dagger$} & \cellcolor[HTML]{FAC791}{0.345} & \cellcolor[HTML]{F59194}{0.227} & \cellcolor[HTML]{F59194}{0.175} & \cellcolor[HTML]{F59194}{0.265$^\dagger$}\\
                    
                    Ours & 
                    \cellcolor[HTML]{F59194}{0.167} & \cellcolor[HTML]{F59194}{0.205} & \cellcolor[HTML]{F59194}{0.212} & \cellcolor[HTML]{F59194} & \cellcolor[HTML]{F59194}{0.093} & \cellcolor[HTML]{F59194}{0.146} & \cellcolor[HTML]{F59194}{0.079} & \cellcolor[HTML]{F59194} & \cellcolor[HTML]{F59194}{0.226} & \cellcolor[HTML]{F59194}{0.232} & \cellcolor[HTML]{F59194}{0.192} & \cellcolor[HTML]{F59194} & \cellcolor[HTML]{F59194}{0.299} & \cellcolor[HTML]{FAC791}{0.275} & \cellcolor[HTML]{FAC791}{0.272} & \cellcolor[HTML]{F59194} \\                 
                    \specialrule{.2em}{.1em}{.1em}
            \end{tabular}
        }
    \end{subtable}

    \begin{subtable}{\linewidth}
        \centering
        \label{tab:dtu_uc}
        \scalebox{0.7}
        {
            \begin{tabular}{lcccccccccccccccc}  
            
                \specialrule{.2em}{.1em}{.1em}
                
                    \multicolumn{1}{c}{Scenes} & 
                    \multicolumn{4}{c}{\textbf{DTU-scan55}} & 
                    \multicolumn{4}{c}{\textbf{DTU-scan63}} & 
                    \multicolumn{4}{c}{\textbf{DTU-scan83}} & 
                    \multicolumn{4}{c}{\textbf{DTU-scan105}} \\
                \cmidrule[0.5pt](rl){2-5}
                \cmidrule[0.5pt](rl){6-9}
                \cmidrule[0.5pt](rl){10-13}
                \cmidrule[0.5pt](rl){14-17}
        
                    \multirow{2}{*}{\diagbox{Methods}{Metrics}} & 
                    \multicolumn{3}{c}{AUSE($\downarrow$)} & 
                    \multirow{2}{*}{CD($\downarrow$)} & 
                    \multicolumn{3}{c}{AUSE($\downarrow$)} & 
                    \multirow{2}{*}{CD($\downarrow$)} & 
                    \multicolumn{3}{c}{AUSE($\downarrow$)} & 
                    \multirow{2}{*}{CD($\downarrow$)} & 
                    \multicolumn{3}{c}{AUSE($\downarrow$)} & 
                    \multirow{2}{*}{CD($\downarrow$)} \\
                \cmidrule[0.3pt](rl){2-4}                       
                \cmidrule[0.3pt](rl){6-8}                        
                \cmidrule[0.3pt](rl){10-12}                     
                \cmidrule[0.3pt](rl){14-16} & 
                    
                    $\Delta$MSE & $\Delta$MAE & 3D &  & 
                    $\Delta$MSE & $\Delta$MAE & 3D &  & 
                    $\Delta$MSE & $\Delta$MAE & 3D &  & 
                    $\Delta$MSE & $\Delta$MAE & 3D &  \\
        
                \toprule
        
                    CFNeRF\textsuperscript{*}  & 
                    0.367 & 0.463 & \diagbox{}{} & 4.205 & 
                    0.385 & 0.426 & \diagbox{}{} & 4.357 & 
                    \cellcolor[HTML]{FAC791}{0.370} & \cellcolor[HTML]{FAC791}{0.465} & \diagbox{}{} & 4.409 & 
                    \cellcolor[HTML]{FAC791}{0.585} & 0.624 & \diagbox{}{} & 3.978 \\
                    
                    ActiveNeRF\textsuperscript{*}  & 
                    0.634 & 0.671 & 0.781 & \cellcolor[HTML]{F59194}{0.630} & 
                    0.512 & 0.502 & 0.923 & \cellcolor[HTML]{FAC791}{1.458} & 
                    1.352 & 1.407 & 1.006 & \cellcolor[HTML]{FAC791}{0.961} & 
                    1.491 & 1.132 & 0.754 & \cellcolor[HTML]{FAC791}{0.849} \\
        
                \hline
        
                    Lee et al.\textsuperscript{*}  & 
                    1.274 & 1.297 & \diagbox{}{} & \cellcolor[HTML]{FAC791} & 
                    1.375 & 1.697 & \diagbox{}{} & \cellcolor[HTML]{F59194} & 
                    1.941 & 1.449 & \diagbox{}{} & \cellcolor[HTML]{F59194} & 
                    2.096 & 1.219 & \diagbox{}{} & \cellcolor[HTML]{F59194} \\
                    
                    Bayes' Rays  & 
                    \cellcolor[HTML]{FAC791}{0.348} & \cellcolor[HTML]{FAC791}{0.290} & \cellcolor[HTML]{FAC791}{0.313} & \cellcolor[HTML]{FAC791}{0.675$^\dagger$} & 
                    \cellcolor[HTML]{FAC791}{0.252} & \cellcolor[HTML]{FAC791}{0.402} & \cellcolor[HTML]{FAC791}{0.499} & \cellcolor[HTML]{F59194}{1.154$^\dagger$} & 
                    0.934 & 0.878 & \cellcolor[HTML]{FAC791}{0.609} & \cellcolor[HTML]{F59194}{1.035$^\dagger$} & 
                    2.136 & 1.440 & \cellcolor[HTML]{FAC791}{0.547} & \cellcolor[HTML]{F59194}{0.839$^\dagger$} \\
                    Ours  & 
                    \cellcolor[HTML]{F59194}{0.231} & \cellcolor[HTML]{F59194}{0.253} & \cellcolor[HTML]{F59194}{0.192} &  \cellcolor[HTML]{FAC791} & 
                    \cellcolor[HTML]{F59194}{0.207} & \cellcolor[HTML]{F59194}{0.256} & \cellcolor[HTML]{F59194}{0.284} &  \cellcolor[HTML]{F59194} & 
                    \cellcolor[HTML]{F59194}{0.232} & \cellcolor[HTML]{F59194}{0.249} & \cellcolor[HTML]{F59194}{0.303} &  \cellcolor[HTML]{F59194} & 
                    \cellcolor[HTML]{F59194}{0.208} & \cellcolor[HTML]{F59194}{0.328} & \cellcolor[HTML]{F59194}{0.363} &  \cellcolor[HTML]{F59194} \\

                \specialrule{.2em}{.1em}{.1em}
                    \multicolumn{1}{c}{Scenes} & 
                    \multicolumn{4}{c}{\textbf{DTU-scan106}} & 
                    \multicolumn{4}{c}{\textbf{DTU-scan114}} & 
                    \multicolumn{4}{c}{\textbf{DTU-scan118}} & 
                    \multicolumn{4}{c}{\textbf{DTU-scan122}} \\
                \cmidrule[0.5pt](rl){2-5}                        
                \cmidrule[0.5pt](rl){6-9}                        
                \cmidrule[0.5pt](rl){10-13}                        
                \cmidrule[0.5pt](rl){14-17}
                    
                    \multirow{2}{*}{\diagbox{Methods}{Metrics}} & 
                    \multicolumn{3}{c}{AUSE($\downarrow$)} & 
                    \multirow{2}{*}{CD($\downarrow$)} & 
                    \multicolumn{3}{c}{AUSE($\downarrow$)} & 
                    \multirow{2}{*}{(CD$\downarrow$)} & 
                    \multicolumn{3}{c}{AUSE($\downarrow$)} & 
                    \multirow{2}{*}{CD($\downarrow$)} & 
                    \multicolumn{3}{c}{AUSE($\downarrow$)} & 
                    \multirow{2}{*}{CD($\downarrow$)} \\
                \cmidrule[0.3pt](rl){2-4}                        
                \cmidrule[0.3pt](rl){6-8}                        
                \cmidrule[0.3pt](rl){10-12}                      
                \cmidrule[0.3pt](rl){14-16} &
                
                    $\Delta$MSE & $\Delta$MAE & 3D &  & 
                    $\Delta$MSE & $\Delta$MAE & 3D &  & 
                    $\Delta$MSE & $\Delta$MAE & 3D &  & 
                    $\Delta$MSE & $\Delta$MAE & 3D &  \\
                    
                \toprule
                
                    CFNeRF\textsuperscript{*}   &
                    0.462 & 0.437 & \diagbox{}{} & 3.530 & 
                    0.366 & 0.419 & \diagbox{}{} & 3.879 & 
                    0.390 & 0.387 & \diagbox{}{} & 3.914 &
                    0.398 & 0.381 & \diagbox{}{} & 3.750 \\
                    
                    ActiveNeRF\textsuperscript{*}  & 
                    1.222 & 1.253 & 1.049 & \cellcolor[HTML]{FAC791}{0.778} &
                    1.489 & 1.167 & 1.160 & \cellcolor[HTML]{FAC791}{0.706} & 
                    0.470 & 0.583 & 1.312 & \cellcolor[HTML]{FAC791}{0.910} & 
                    0.438 & 0.551 & 0.923 & \cellcolor[HTML]{FAC791}{0.907} \\
                    
                \hline
                    Lee et al.\textsuperscript{*}  & 
                    2.177 & 1.515 & \diagbox{}{} & \cellcolor[HTML]{F59194} & 
                    1.898 & 1.633 & \diagbox{}{} & \cellcolor[HTML]{F59194} & 
                    1.341 & 1.504 & \diagbox{}{} & \cellcolor[HTML]{F59194} & 
                    1.846 & 1.816 & \diagbox{}{} & \cellcolor[HTML]{F59194} \\
                    
                    Bayes' Rays  & 
                    \cellcolor[HTML]{FAC791}{0.181} & \cellcolor[HTML]{FAC791}{0.246} & \cellcolor[HTML]{FAC791}{0.356} & \cellcolor[HTML]{F59194}{0.656$^\dagger$} & 
                    \cellcolor[HTML]{FAC791}{0.204} & \cellcolor[HTML]{FAC791}{0.341} & \cellcolor[HTML]{FAC791}{0.312} & \cellcolor[HTML]{F59194}{0.519$^\dagger$} &
                    \cellcolor[HTML]{FAC791}{0.267} & \cellcolor[HTML]{FAC791}{0.355} & \cellcolor[HTML]{FAC791}{0.393} & \cellcolor[HTML]{F59194}{0.824$^\dagger$} & \cellcolor[HTML]{FAC791}{0.330} & \cellcolor[HTML]{FAC791}{0.291} & \cellcolor[HTML]{FAC791}{0.290} & \cellcolor[HTML]{F59194}{0.864$^\dagger$}\\
                    
                    Ours & 
                    \cellcolor[HTML]{F59194}{0.150} & \cellcolor[HTML]{F59194}{0.204} & \cellcolor[HTML]{F59194}{0.213} &  \cellcolor[HTML]{F59194} & \cellcolor[HTML]{F59194}{0.171} & \cellcolor[HTML]{F59194}{0.319} & \cellcolor[HTML]{F59194}{0.188} & \cellcolor[HTML]{F59194} & \cellcolor[HTML]{F59194}{0.217} & \cellcolor[HTML]{F59194}{0.188} & \cellcolor[HTML]{F59194}{0.214} & \cellcolor[HTML]{F59194} & \cellcolor[HTML]{F59194}{0.233} & \cellcolor[HTML]{F59194}{0.252} & \cellcolor[HTML]{F59194}{0.205} &  \cellcolor[HTML]{F59194} \\                 
                    \specialrule{.2em}{.1em}{.1em}
            \end{tabular}
        }
    \end{subtable}
    \caption{ \textbf{Uncertainty Quantification for the BlendedMVS and DTU datasets}. Best results are highlighted as \colorbox[HTML]{F59194}{\textbf{first}},\colorbox[HTML]{FAC791}{\textbf{second}}. $\dagger$~Bayes’ Rays and Unc-NeRF evaluate our trained model as post-hoc frameworks and share the same CD metric.} 
    \label{tab:dtu_bmvs_uc}
\end{table*}

\begin{figure*}[!ht] 
\centering
\scriptsize
    \includegraphics[width=\linewidth]{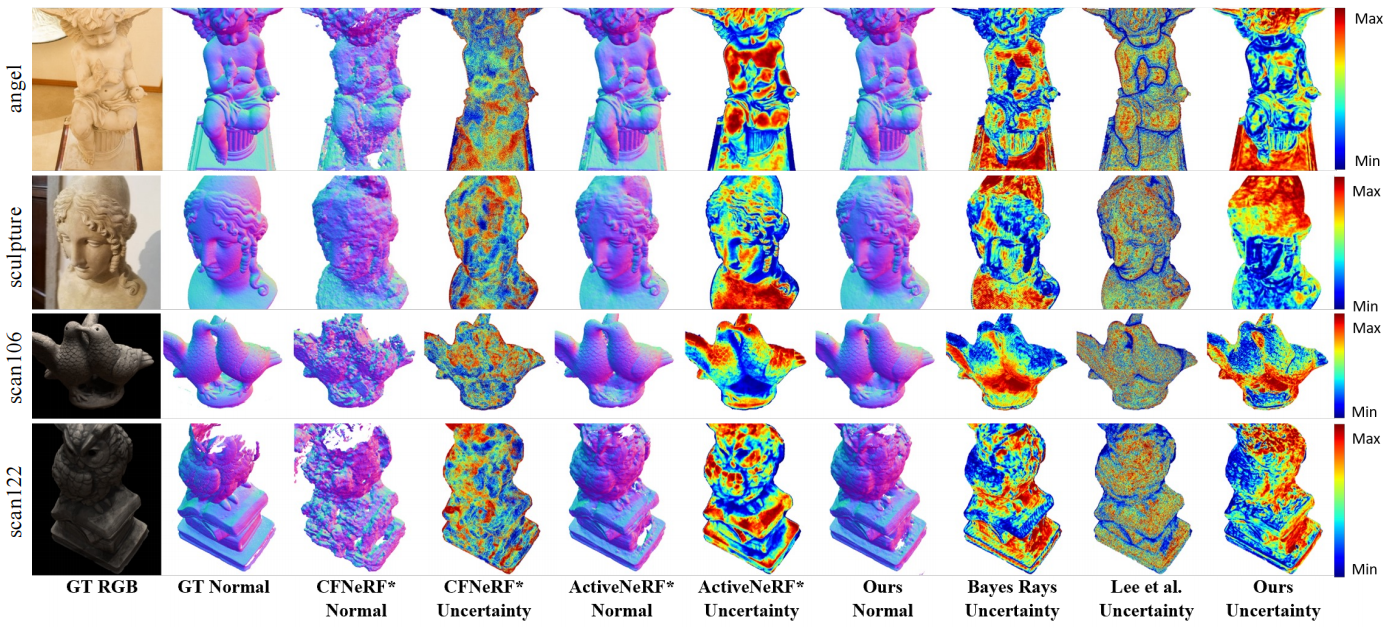}
    \caption{\textbf{Qualitative comparison of the uncertainty estimation with 2D depth.}
    Our estimated uncertainties are more closely aligned with the error between the GT and predicted geometry than other baselines~\cite{pan2022activenerf,shen2022conditional,goli2023bayes,lee2022uncertainty}. Following \cite{goli2023bayes}, uncertainties are colored based on the ranking order of uncertainty scores.}
    \label{fig:uncertainty_quantification}
\end{figure*}

\section{Experiment}
In this section, we first assess the efficacy of GURecon in uncertainty quantification. 
Then we perform ablation studies to validate each component within our framework, demonstrating its versatility across different numbers of training images and various neural surface models. 
Lastly, we demonstrate our plug-and-play capability of GURecon by applying it to the task of incremental reconstruction and comparing it against other NeRF-based NBV selection methods.

\begin{table}[t]
    \centering
    \label{tab:ablation}
        \begin{subtable}{0.475 \textwidth}
            \centering
            \setlength{\tabcolsep}{4.5mm}
            \scalebox{0.65}
            {\large
            \begin{tabular}{l|ccc}
                \specialrule{.2em}{.1em}{.1em}
                    Scheme & $\mathrm{AUSE}_\mathrm{MSE}$ & $\mathrm{AUSE}_\mathrm{MAE}$ & $\mathrm{AUSE}_\mathrm{3D}$ \\
                    \hline
                    w/o decouple finetune & 0.104 & 0.137 & 0.125 \\
                    with pixel-based consistency & 0.675 & 0.713 & 0.792 \\
                    with smaller patch size 7 & 0.135 & 0.254 & 0.179 \\
                    with larger patch size 15 & 0.207 & 0.195 & 0.214 \\
                    Full Model & \textbf{0.058} & \textbf{0.094} & \textbf{0.091} \\
                \specialrule{.2em}{.1em}{.1em}
            \end{tabular}
            }
            \label{subtab:ablation_module}
        \end{subtable}
    \caption{Ablation studies on BlendedMVS dataset.}
    \label{tab:ablation}
\end{table}

\subsection{Uncertainty Quantification}
\label{ssec:uncern_quantification}
\boldtitle{Datasets} 
We evaluate our method over three widely used benchmark datasets: the DTU dataset \cite{jensen2014large}, the BlendedMVS dataset \cite{yao2020blendedmvs}, and the Tank and Template~(TNT) dataset \cite{knapitsch2017tanks}. 
These datasets offer calibrated multi-view images, along with object masks and high-fidelity 3D models serving as the ground truth.
The DTU dataset comprises object scans, with each scene containing 49 or 64 views from concentrated perspectives.
We selected eight scenes with diverse materials, all exposed to challenging ambient lighting conditions.
The BlendedMVS dataset contains a large collection of indoor and outdoor scenes, each featuring 360-degree surround view captures with varying scales and numbers of images.
Additionally, we conduct experiments on 3 large-scale outdoor scenes from the TNT dataset with more randomized viewpoints and biased captures.
As discussed in \cite{shen2022conditional}, training with sparse images can ensure variations in the reconstruction quality, providing an ideal setup to evaluate the uncertainty modeling ability. 
Therefore, based on the spatial distribution within each scene, we uniformly sample a sparse number of views for the training ($\sim$6) and test ($\sim$3) sets in the DTU dataset, for the BlendedMVS and TNT datasets, we uniformly sample 25\% images for the training set and choose 4 adjacent images as the test set.

\boldtitle{Metrics} 
In the experiment for uncertainty quantification, following previous methods~\cite{shen2022conditional,goli2023bayes},  
we calculate the Area Under Sparsification Error~(AUSE), a widely used metric to assess the quality of model uncertainty. Given the predicted depth error and predicted uncertainty of each pixel in the test image, we gradually remove the top $t\% (t=1\sim100)$ pixels according to two criteria: once based on GT depth error, once based on predicted uncertainty, and compute the average depth error for the remaining pixels. The area between the curves obtained from the two criteria is the AUSE, which reflects the correlation between predicted uncertainty and actual depth error. 
In addition to calculating AUSE based on the Mean Absolute depth Error~($\Delta$MAE) and Mean Squared Error~($\Delta$MSE), we also compute the AUSE based on the closest distance of each point to the ground truth geometry as the 3D geometric error, \ie, $\mathrm{AUSE}_\mathrm{3D}$, which reflects geometric uncertainty from a 3D perspective.
Finally, we evaluate the accuracy of surfaces reconstructed by different methods using the Chamfer Distance~(CD). 
Since the scales among scenes are inconsistent in both BlendedMVS and TNT datasets, we uniformly normalize the scenes to fit within the unit sphere to compute geometric metrics. 

\boldtitle{Baselines}
We compare ours with previous works designed for uncertainty estimation in NeRF: CFNeRF~\cite{shen2022conditional}, ActiveNeRF~\cite{pan2022activenerf}, Bayes' Rays~\cite{goli2023bayes}, and Uncertainty-Guided NeRF~\cite{lee2022uncertainty}.
Although the ability to model uncertainty should be independent of reconstruction quality, considering they are all designed based on NeRF~\cite{mildenhall2021nerf}, while ours is designed for the SDF backend, to avoid differences in geometric errors caused by representation affecting the assessment of uncertainty modeling capabilities, we make structural modifications to each method (labeled with \textsuperscript{*}), making them applicable to neural surface representation.
Please refer to the Supp. Mat. for details.
To be noted, for Bayes' Rays~\cite{goli2023bayes} and Uncertainty-Guided NeRF~\cite{lee2022uncertainty}, 
we directly migrate them as post-hoc frameworks to evaluate the model $(\SDFNet(\NetParams_{s}),\ColorNet(\NetParams_c))$ trained by our method.
From the geometric perspective, although ActiveNeRF\textsuperscript{*} and Bayes' Rays cannot directly model the uncertainty of 3D points since they rely on the pixel-level volumetric rendering process, 
we feed the surface points into their uncertainty model as a rough estimation and compare them in terms of $\mathrm{AUSE}_\mathrm{3D}$ as a reference for better understanding.

\begin{figure}[t] 
    \centering
    \includegraphics[width=1.0 \linewidth]{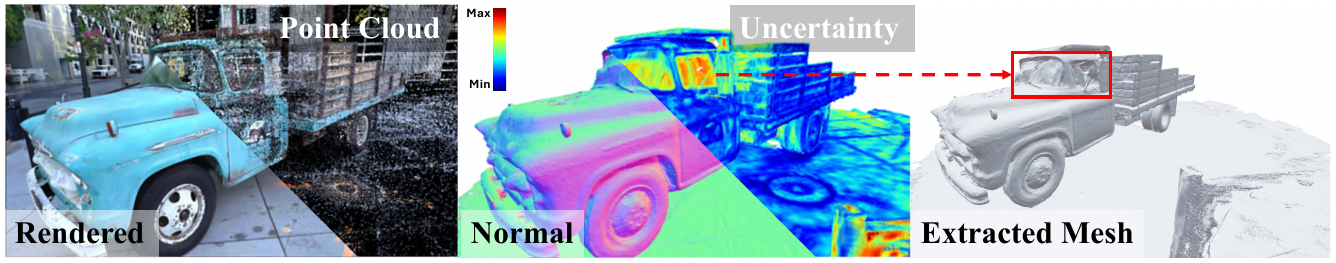}
    \caption{\textbf{Extension to 2DGS.} Our uncertainty distillation can be migrated to various neural surface representations. We show the extension to 2DGS~\cite{huang20242d}.}
    \label{fig:GS_uncertainty}
\end{figure}

\begin{figure*}[t] \centering
    \includegraphics[width=0.95\linewidth]{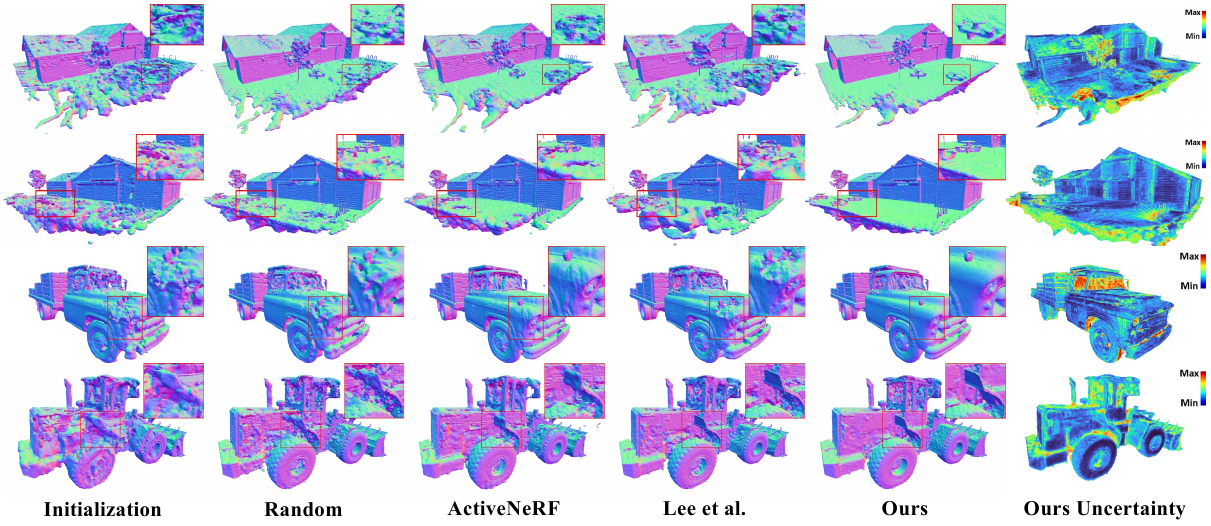}
    \caption{\textbf{Incremental results on TNT dataset}. 
    Our scheme reconstructs more details while ensuring a smoother surface.}
    \label{fig:Incremental}
\end{figure*}

\boldtitle{Results}
As illustrated in Fig.~\ref{fig:uncertainty_quantification} and Table~\ref{tab:dtu_bmvs_uc}, under sparse-view setting, 
CFNeRF\textsuperscript{*} exhibits severe degradation in reconstruction quality and inadequate predictive capabilities for uncertainty, particularly evident in the DTU datasets where training is conducted with only six views. 
Considering the uncertainty prediction of ActiveNeRF\textsuperscript{*} is based on modeling the predicted RGB, which is vulnerable to significant disruption when substantial color variations exist across different viewpoints, it does not adequately reflect geometric uncertainty.
As NeuS~\cite{wang2021neus} is designed under the assumption of an ideal impulse function distribution, the estimation of uncertainty based on the entropy of the weight distribution on sampled rays in Uncertainty-Guided NeRF\textsuperscript{*} becomes ineffective.
Bayes' Rays has a relatively reasonable performance of modeling uncertainty among existing approaches. However, it still models the uncertainty from the perspective of RGB rendering by perturbing sampling points and tends to predict high uncertainty for regions with abundant repetitive textures regardless of the reconstruction, such as the case of Angel and Sculpture shown in Fig.~\ref{fig:uncertainty_quantification}.
Compared to other methods, as our approach leverages the advantages of multi-view geometric consistency and decoupling view-dependent factors, GURecon achieves a significant improvement. Even facing scenes with texture repetitions, lighting interferences, and sparse training views as shown in Fig.~\ref{fig:uncertainty_quantification}, our method accurately distinguishes unreliable regions and quantitatively evaluates the uncertainty. As the AUSE curves shown in Fig.~\ref{fig:ause_curve}, the depth error identified by ours closely approximates the actual variation curve. 
From Fig.~\ref{fig:ours_res} and Fig.~\ref{fig:ours&bayes}, we can also see that our learned 3D geometric uncertainties align well with the real 3D geometric error in both indoor and outdoor large-scale scenes.

\subsection{Ablation Study}
We conduct an ablation study to demonstrate the effectiveness of each module in the proposed method.

\boldtitle{Fine-tuning with Decoupling Modules} 
As shown in the Fig.~\ref{fig:finetune} and Table~\ref{tab:ablation}, fine-tuning with decoupling modules effectively addresses the misclassification issue wherein regions with good reconstruction quality are erroneously classified as reconstruction failures due to lighting interference in the calculation of geometric consistency. 

\boldtitle{Different Sizes of Patches in Consistency} 
As shown in Table~\ref{tab:ablation}, the utilization of pixel-based and small patch-based consistency fails to accurately reflect geometric uncertainty due to their sensitivity to view-dependent factors such as lighting. Large patches fail to capture geometric consistency in detailed areas such as edges and corners. The patch size used is 11x11.
Please refer to more ablation in Supp. Mat.

\boldtitle{Plug-and-play extension to 2DGS} 
Our proposed uncertainty distillation can be migrated to various neural surface representations. We extend it to the latest surface reconstruction work 2DGS~\cite{huang20242d}. 
We utilize the GS corresponding to the median depth of each pixel as the intersection with the surface and employ the direction of its shortest axis as the normal for homography warping. With the proposed distillation method, we can supervise an additional attribute of uncertainty for the GS located on the surface, as shown in Fig.~\ref{fig:GS_uncertainty}. Please refer to more details in Supp. Mat.

\renewcommand{\arraystretch}{1}
\begin{table}[t] 
\centering
\begin{minipage}[t]{0.975\linewidth}
    \setlength{\tabcolsep}{5pt}
    \resizebox{\textwidth}{!}
    {  \Huge
        \begin{tabular}{lcccccc}   
            \specialrule{.2em}{.1em}{.1em}
            \multirow{2}{*}{\diagbox{Methods}{Scenes}} & \multicolumn{2}{c}{\textbf{Barn}} & \multicolumn{2}{c}{\textbf{Caterpillar}} & \multicolumn{2}{c}{\textbf{Truck}} \\
            \cmidrule[0.5pt](rl){2-3} \cmidrule[0.5pt](rl){4-5} \cmidrule[0.5pt](rl){6-7}
                                    & \multicolumn{1}{c}{$\mathrm{CD}$($\downarrow$)}        & \multicolumn{1}{c}{$\mathrm{PSNR}$($\uparrow$)}                      & \multicolumn{1}{c}{$\mathrm{CD}$($\downarrow$)}        & \multicolumn{1}{c}{$\mathrm{PSNR}$($\uparrow$)}             & \multicolumn{1}{c}{$\mathrm{CD}$($\downarrow$)}        & \multicolumn{1}{c}{$\mathrm{PSNR}$($\uparrow$)}         \\
            \toprule
              \textbf{Random}          & \multicolumn{1}{c}{1.033} & \multicolumn{1}{c}{22.69} & \multicolumn{1}{c}{0.809} &\multicolumn{1}{c}{19.71} &\multicolumn{1}{c}{2.248} & \multicolumn{1}{c}{20.89}\\
            \textbf{ActiveNeRF* \cite{pan2022activenerf}  }        &    \multicolumn{1}{c}{1.002} & \multicolumn{1}{c}{22.72} & \multicolumn{1}{c}{0.733} & \multicolumn{1}{c}{19.24} & \multicolumn{1}{c}{2.123} & \multicolumn{1}{c}{20.73} \\
           \textbf{Lee et al.* \cite{lee2022uncertainty} }   &    \multicolumn{1}{c} {0.983} &  \multicolumn{1}{c}{22.02} &  \multicolumn{1}{c}{0.778} &  \multicolumn{1}{c}{19.24} &  \multicolumn{1}{c}{2.212} &  \multicolumn{1}{c}{20.40} \\
           \textbf{Ours}  &    \multicolumn{1}{c}{\bf{0.947}} &  \multicolumn{1}{c}{\bf{23.21}} &  \multicolumn{1}{c}{\bf{0.705}} &  \multicolumn{1}{c}{\bf{20.06}} &  \multicolumn{1}{c}{\bf{2.059}} &  \multicolumn{1}{c}{\bf{21.49}}\\
            \specialrule{.2em}{.1em}{.1em}
        \end{tabular}
    }
\end{minipage}
 \caption{\textbf{Quantitative comparison of NBV strategies on TNT dataset.} The best results are highlighted in \bf{bold}.} 
\label{tab:incre}
\end{table}

\subsection{Evaluation on Incremental Reconstruction}
\label{ssec:increRecon}
\boldtitle{Datasets and Metrics}
We select the same large-scale scenarios from the TNT dataset as used in Sec.~\ref{ssec:uncern_quantification} to measure the effectiveness of the incremental reconstruction strategy. We report Chamfer Distance for surface evaluation and peak signal-to-noise ratio (PSNR) for image synthesis qualities. 

\boldtitle{Baselines}
We compare ours with two representative NeRF-based NBV methods: Uncertainty-Guided NeRF~\cite{lee2022uncertainty} and ActiveNeRF~\cite{pan2022activenerf} discussed in Sec.~\ref{ssec:uncern_quantification}, alongside a completely randomized NBV strategy.
Considering the substantial differences in geometric quality between NeRF and NeuS representations, we adopt the same strategy described in Sec.~\ref{ssec:uncern_quantification}.

\boldtitle{Implementation Details}
We follow the initialization strategy in \cite{lee2022uncertainty} to divide the space into several regions uniformly and select one viewpoint from each region for both the initial training set and the test set. During each selection for Next Best View (NBV), we assess the uncertainty of the remaining viewpoints in each region and select the one with the highest score to augment the training set. As our method directly models the uncertainty of surface points, we utilize sphere tracing for root-finding and achieve rapid surface rendering of uncertainty for new viewpoints.

\boldtitle{Results} 
As shown in Fig.~\ref{fig:Incremental} and Table~\ref{tab:incre}, the NBV strategy with our geometric uncertainty achieves the best reconstruction results under the same limited number of views (30\% of the total image). Compared to other methods, our approach reconstructs more details while ensuring a smoother surface. Please refer to Supp. Mat. for a more qualitative comparison.

\section{Conclusion}

In this paper, we introduce GURecon, a novel approach for learning a 3D geometric uncertainty field for neural surface models. Unlike existing methods that model rendering-based pixel-wise uncertainty, the proposed GURecon exploits the multi-view consistency to accurately model the geometric uncertainty. Moreover, additional decoupled fields are learned for robust uncertainty estimation. 
Comprehensive experiments have demonstrated our superior performance compared to existing methods.
While our approach works well in small-scale textureless regions, its performance is limited in extreme scenarios with large textureless areas (\eg, white walls), where high-level semantic information can be incorporated in future work.

\clearpage
\section*{Acknowledgments}
We express our gratitude to all the anonymous reviewers for their professional and
constructive comments. This work was partially supported by the NSFC (No.~62441222), Information Technology Center and State Key Lab of CAD\&CG, Zhejiang University.

\bibliography{aaai25}

\begin{thebibliography}{35}
\providecommand{\natexlab}[1]{#1}

\bibitem[{Campbell et~al.(2008)Campbell, Vogiatzis, Hern{\'a}ndez, and Cipolla}]{campbell2008using}
Campbell, N.~D.; Vogiatzis, G.; Hern{\'a}ndez, C.; and Cipolla, R. 2008.
\newblock Using multiple hypotheses to improve depth-maps for multi-view stereo.
\newblock In \emph{Eur. Conf. Comput. Vis.}, 766--779. Springer.

\bibitem[{Chen et~al.(2023)Chen, Chen, Wang, and Pollefeys}]{chen2023leveraging}
Chen, L.; Chen, W.; Wang, R.; and Pollefeys, M. 2023.
\newblock Leveraging Neural Radiance Fields for Uncertainty-Aware Visual Localization.
\newblock \emph{arXiv preprint arXiv:2310.06984}.

\bibitem[{Darmon et~al.(2022)Darmon, Bascle, Devaux, Monasse, and Aubry}]{darmon2022improving}
Darmon, F.; Bascle, B.; Devaux, J.-C.; Monasse, P.; and Aubry, M. 2022.
\newblock Improving neural implicit surfaces geometry with patch warping.
\newblock In \emph{IEEE Conf. Comput. Vis. Pattern Recog.}, 6260--6269.

\bibitem[{Fan et~al.(2023)Fan, Skorokhodov, Voynov, Ignatyev, Burnaev, Wonka, and Wang}]{fan2023factored}
Fan, Y.; Skorokhodov, I.; Voynov, O.; Ignatyev, S.; Burnaev, E.; Wonka, P.; and Wang, Y. 2023.
\newblock Factored-NeuS: Reconstructing Surfaces, Illumination, and Materials of Possibly Glossy Objects.
\newblock \emph{arXiv preprint arXiv:2305.17929}.

\bibitem[{Fu et~al.(2022)Fu, Xu, Ong, and Tao}]{fu2022geo}
Fu, Q.; Xu, Q.; Ong, Y.~S.; and Tao, W. 2022.
\newblock Geo-neus: Geometry-consistent neural implicit surfaces learning for multi-view reconstruction.
\newblock \emph{Adv. Neural Inform. Process. Syst.}, 35: 3403--3416.

\bibitem[{Ge et~al.(2023)Ge, Hu, Zhao, Liu, and Chen}]{ge2023ref}
Ge, W.; Hu, T.; Zhao, H.; Liu, S.; and Chen, Y.-C. 2023.
\newblock Ref-NeuS: Ambiguity-Reduced Neural Implicit Surface Learning for Multi-View Reconstruction with Reflection.
\newblock \emph{arXiv preprint arXiv:2303.10840}.

\bibitem[{Goli et~al.(2023)Goli, Reading, Selll{\'a}n, Jacobson, and Tagliasacchi}]{goli2023bayes}
Goli, L.; Reading, C.; Selll{\'a}n, S.; Jacobson, A.; and Tagliasacchi, A. 2023.
\newblock Bayes' Rays: Uncertainty Quantification for Neural Radiance Fields.
\newblock \emph{arXiv preprint arXiv:2309.03185}.

\bibitem[{Hu and Mordohai(2012)}]{hu2012quantitative}
Hu, X.; and Mordohai, P. 2012.
\newblock A Quantitative Evaluation of Confidence Measures for Stereo Vision.
\newblock \emph{IEEE Transactions on Pattern Analysis and Machine Intelligence}, 34(11): 2121--2133.

\bibitem[{Huang et~al.(2024)Huang, Yu, Chen, Geiger, and Gao}]{huang20242d}
Huang, B.; Yu, Z.; Chen, A.; Geiger, A.; and Gao, S. 2024.
\newblock 2d gaussian splatting for geometrically accurate radiance fields.
\newblock In \emph{ACM SIGGRAPH 2024 Conference Papers}, 1--11.

\bibitem[{Jensen et~al.(2014)Jensen, Dahl, Vogiatzis, Tola, and Aan{\ae}s}]{jensen2014large}
Jensen, R.; Dahl, A.; Vogiatzis, G.; Tola, E.; and Aan{\ae}s, H. 2014.
\newblock Large scale multi-view stereopsis evaluation.
\newblock In \emph{IEEE Conf. Comput. Vis. Pattern Recog.}, 406--413.

\bibitem[{Jiang, Lei, and Daniilidis(2023)}]{jiang2023fisherrf}
Jiang, W.; Lei, B.; and Daniilidis, K. 2023.
\newblock FisherRF: Active View Selection and Uncertainty Quantification for Radiance Fields using Fisher Information.
\newblock \emph{arXiv preprint arXiv:2311.17874}.

\bibitem[{Knapitsch et~al.(2017)Knapitsch, Park, Zhou, and Koltun}]{knapitsch2017tanks}
Knapitsch, A.; Park, J.; Zhou, Q.-Y.; and Koltun, V. 2017.
\newblock Tanks and temples: Benchmarking large-scale scene reconstruction.
\newblock \emph{ACM Trans. Graph.}, 36(4): 1--13.

\bibitem[{Lee et~al.(2022)Lee, Chen, Wang, Liniger, Kumar, and Yu}]{lee2022uncertainty}
Lee, S.; Chen, L.; Wang, J.; Liniger, A.; Kumar, S.; and Yu, F. 2022.
\newblock Uncertainty guided policy for active robotic 3d reconstruction using neural radiance fields.
\newblock \emph{IEEE Robotics and Automation Letters}, 7(4): 12070--12077.

\bibitem[{Liu et~al.(2020)Liu, Gu, Zaw~Lin, Chua, and Theobalt}]{liu2020neural}
Liu, L.; Gu, J.; Zaw~Lin, K.; Chua, T.-S.; and Theobalt, C. 2020.
\newblock Neural sparse voxel fields.
\newblock \emph{Adv. Neural Inform. Process. Syst.}, 33: 15651--15663.

\bibitem[{Martin-Brualla et~al.(2021)Martin-Brualla, Radwan, Sajjadi, Barron, Dosovitskiy, and Duckworth}]{martin2021nerf}
Martin-Brualla, R.; Radwan, N.; Sajjadi, M.~S.; Barron, J.~T.; Dosovitskiy, A.; and Duckworth, D. 2021.
\newblock Nerf in the wild: Neural radiance fields for unconstrained photo collections.
\newblock In \emph{IEEE Conf. Comput. Vis. Pattern Recog.}, 7210--7219.

\bibitem[{Mildenhall et~al.(2021)Mildenhall, Srinivasan, Tancik, Barron, Ramamoorthi, and Ng}]{mildenhall2021nerf}
Mildenhall, B.; Srinivasan, P.~P.; Tancik, M.; Barron, J.~T.; Ramamoorthi, R.; and Ng, R. 2021.
\newblock Nerf: Representing scenes as neural radiance fields for view synthesis.
\newblock \emph{Communications of the ACM}, 65(1): 99--106.

\bibitem[{Oechsle, Peng, and Geiger(2021)}]{oechsle2021unisurf}
Oechsle, M.; Peng, S.; and Geiger, A. 2021.
\newblock Unisurf: Unifying neural implicit surfaces and radiance fields for multi-view reconstruction.
\newblock In \emph{Int. Conf. Comput. Vis.}, 5589--5599.

\bibitem[{Pan et~al.(2022)Pan, Lai, Song, and Huang}]{pan2022activenerf}
Pan, X.; Lai, Z.; Song, S.; and Huang, G. 2022.
\newblock Activenerf: Learning where to see with uncertainty estimation.
\newblock In \emph{Eur. Conf. Comput. Vis.}, 230--246.

\bibitem[{Ran et~al.(2023)Ran, Zeng, He, Chen, Li, Chen, Lee, and Ye}]{ran2023neurar}
Ran, Y.; Zeng, J.; He, S.; Chen, J.; Li, L.; Chen, Y.; Lee, G.; and Ye, Q. 2023.
\newblock NeurAR: Neural Uncertainty for Autonomous 3D Reconstruction With Implicit Neural Representations.
\newblock \emph{IEEE Robotics and Automation Letters}, 8(2): 1125--1132.

\bibitem[{Sch{\"o}nberger et~al.(2016)Sch{\"o}nberger, Zheng, Frahm, and Pollefeys}]{schonberger2016pixelwise}
Sch{\"o}nberger, J.~L.; Zheng, E.; Frahm, J.-M.; and Pollefeys, M. 2016.
\newblock Pixelwise view selection for unstructured multi-view stereo.
\newblock In \emph{Eur. Conf. Comput. Vis.}, 501--518. Springer.

\bibitem[{Shen et~al.(2022)Shen, Agudo, Moreno-Noguer, and Ruiz}]{shen2022conditional}
Shen, J.; Agudo, A.; Moreno-Noguer, F.; and Ruiz, A. 2022.
\newblock Conditional-flow NeRF: Accurate 3D modelling with reliable uncertainty quantification.
\newblock In \emph{Eur. Conf. Comput. Vis.}, 540--557. Springer.

\bibitem[{Shen et~al.(2021)Shen, Ruiz, Agudo, and Moreno-Noguer}]{shen2021stochastic}
Shen, J.; Ruiz, A.; Agudo, A.; and Moreno-Noguer, F. 2021.
\newblock Stochastic neural radiance fields: Quantifying uncertainty in implicit 3d representations.
\newblock In \emph{International Conference on 3D Vision (3DV)}, 972--981. IEEE.

\bibitem[{Stereopsis(2010)}]{stereopsis2010accurate}
Stereopsis, R.~M. 2010.
\newblock Accurate, Dense, and Robust Multiview Stereopsis.
\newblock \emph{IEEE Trans. Pattern Anal. Mach. Intell.}, 32(8).

\bibitem[{S{\"u}nderhauf, Abou-Chakra, and Miller(2023)}]{sunderhauf2023density}
S{\"u}nderhauf, N.; Abou-Chakra, J.; and Miller, D. 2023.
\newblock Density-aware nerf ensembles: Quantifying predictive uncertainty in neural radiance fields.
\newblock 9370--9376. IEEE.

\bibitem[{Tang et~al.(2023)Tang, Zhou, Chen, Hu, Ding, Wang, and Zeng}]{tang2023delicate}
Tang, J.; Zhou, H.; Chen, X.; Hu, T.; Ding, E.; Wang, J.; and Zeng, G. 2023.
\newblock Delicate textured mesh recovery from nerf via adaptive surface refinement.
\newblock \emph{arXiv preprint arXiv:2303.02091}.

\bibitem[{Verbin et~al.(2022)Verbin, Hedman, Mildenhall, Zickler, Barron, and Srinivasan}]{verbin2022refnerf}
Verbin, D.; Hedman, P.; Mildenhall, B.; Zickler, T.; Barron, J.~T.; and Srinivasan, P.~P. 2022.
\newblock {Ref-NeRF}: Structured View-Dependent Appearance for Neural Radiance Fields.
\newblock In \emph{IEEE Conf. Comput. Vis. Pattern Recog.}

\bibitem[{Wang et~al.(2022)Wang, Wang, Long, Theobalt, Komura, Liu, and Wang}]{wang2022neuris}
Wang, J.; Wang, P.; Long, X.; Theobalt, C.; Komura, T.; Liu, L.; and Wang, W. 2022.
\newblock Neuris: Neural reconstruction of indoor scenes using normal priors.
\newblock In \emph{European Conference on Computer Vision}, 139--155. Springer.

\bibitem[{Wang et~al.(2021)Wang, Liu, Liu, Theobalt, Komura, and Wang}]{wang2021neus}
Wang, P.; Liu, L.; Liu, Y.; Theobalt, C.; Komura, T.; and Wang, W. 2021.
\newblock Neus: Learning neural implicit surfaces by volume rendering for multi-view reconstruction.
\newblock \emph{Adv. Neural Inform. Process. Syst.}

\bibitem[{Xiao et~al.(2024)Xiao, Xu, Yu, and Gao}]{xiao2024debsdf}
Xiao, Y.; Xu, J.; Yu, Z.; and Gao, S. 2024.
\newblock Debsdf: Delving into the details and bias of neural indoor scene reconstruction.
\newblock \emph{IEEE Transactions on Pattern Analysis and Machine Intelligence}.

\bibitem[{Yao et~al.(2020)Yao, Luo, Li, Zhang, Ren, Zhou, Fang, and Quan}]{yao2020blendedmvs}
Yao, Y.; Luo, Z.; Li, S.; Zhang, J.; Ren, Y.; Zhou, L.; Fang, T.; and Quan, L. 2020.
\newblock Blendedmvs: A large-scale dataset for generalized multi-view stereo networks.
\newblock In \emph{IEEE Conf. Comput. Vis. Pattern Recog.}, 1790--1799.

\bibitem[{Yariv et~al.(2021)Yariv, Gu, Kasten, and Lipman}]{yariv2021volume}
Yariv, L.; Gu, J.; Kasten, Y.; and Lipman, Y. 2021.
\newblock Volume rendering of neural implicit surfaces.
\newblock \emph{Adv. Neural Inform. Process. Syst.}, 34: 4805--4815.

\bibitem[{Yu et~al.(2022)Yu, Peng, Niemeyer, Sattler, and Geiger}]{yu2022monosdf}
Yu, Z.; Peng, S.; Niemeyer, M.; Sattler, T.; and Geiger, A. 2022.
\newblock Monosdf: Exploring monocular geometric cues for neural implicit surface reconstruction.
\newblock \emph{Adv. Neural Inform. Process. Syst.}, 35: 25018--25032.

\bibitem[{Zhan et~al.(2022)Zhan, Zheng, Xu, Reid, and Rezatofighi}]{zhan2022activermap}
Zhan, H.; Zheng, J.; Xu, Y.; Reid, I.; and Rezatofighi, H. 2022.
\newblock ActiveRMAP: Radiance Field for Active Mapping And Planning.
\newblock \emph{arXiv preprint arXiv:2211.12656}.

\bibitem[{Zhang et~al.(2020)Zhang, Riegler, Snavely, and Koltun}]{zhang2020nerf++}
Zhang, K.; Riegler, G.; Snavely, N.; and Koltun, V. 2020.
\newblock Nerf++: Analyzing and improving neural radiance fields.
\newblock \emph{arXiv preprint arXiv:2010.07492}.

\bibitem[{Zhao et~al.(2022)Zhao, Jiang, Yao, Zhang, Wang, Dai, Zhong, Zhang, Wu, Xu et~al.}]{zhao2022human}
Zhao, F.; Jiang, Y.; Yao, K.; Zhang, J.; Wang, L.; Dai, H.; Zhong, Y.; Zhang, Y.; Wu, M.; Xu, L.; et~al. 2022.
\newblock Human performance modeling and rendering via neural animated mesh.
\newblock \emph{ACM Trans. Graph.}, 41(6): 1--17.

\end{thebibliography}
\end{document}